# Spatio-temporal Aware Non-negative Component Representation for Action Recognition

Jianhong Wang,[1] Tian Lan,[2] Xu Zhang,[1] Limin Luo[1]
[1] Laboratory of Image Sciences and Technology, Southeast University, Nanjing, 210096, China
[2] Ginkgo LLC, Incline Village, NV, 89451, USA

## ABSTRACT
This paper presents a novel mid-level representation for action recognition, named spatio-temporal aware non-negative component representation (STANNCR). The proposed STANNCR is based on action component and incorporates the spatial-temporal information. We first introduce a spatial-temporal distribution vector (STDV) to model the distributions of local feature locations in a compact and discriminative manner. Then we employ non-negative matrix factorization (NMF) to learn the action components and encode the video samples. The action component considers the correlations of visual words, which effectively bridge the sematic gap in action recognition. To incorporate the spatial-temporal cues for final representation, the STDV is used as the part of graph regularization for NMF. The fusion of spatial-temporal information makes the STANNCR more discriminative, and our fusion manner is more compact than traditional method of concatenating vectors. The proposed approach is extensively evaluated on three public datasets. The experimental results demonstrate the effectiveness of STANNCR for action recognition.

## 1. INTRODUCTION

Human action recognition in videos is one of most active research topics in the field of computer vision and pattern recognition. It has a wide range of applications such as video content analysis, video retrieval, surveillance event detection and human-computer interaction (HCI) [1]. However, it is still a challenging topic due to the significant intra-class variations, clutter, occlusion and other fundamental difficulties [2].

The key problem for action recognition is how to represent different action video clips effectively and discriminately. Local features with Bag of Visual Words (BoVW) are most popular framework for representation in most of recent action recognition approaches. In this framework, features are encoded with the visual words in codebook and a histogram of word occurrences is used to represent a video. A significant progress has been made in the development of local features, such as HOG/HOF [3], HOG3D [4], Gist3D [5], and dense trajectory [6]. Meanwhile, A number of encoding method have also been proposed in image and video recognition, e.g., local soft assignment [7], sparse coding [8], and locality-constrained linear coding [9]. These approaches reduce information loss by relaxing the restrictive cardinality constraint in coding features.

While impressive progress has been made, there are still some problems in BoVW framework that need to be addressed. First, the BoVW representation only contains statistics of unordered visual words, the inside relationship between different visual words have not been considered. Meanwhile, the visual words in the codebook do not have any explicit semantics, which limits the discrimination ability of BoVW framework. Another drawback is that the BoVW representation ignores the information concerning the spatial-temporal locations of local features. Obviously, the spatial-temporal locations and distribution may convey useful cue for action recognition, however as the many unconstrained factors in real world videos, it's not easy to use location information directly. The failure of capturing spatio-temporal location information leads to a relatively worse classification accuracy for action recognition.

To simultaneously solve these problems, in this paper, we propose a novel representation for action recognition, named Spatial-temporal Aware Non-Negative Component Representation (STANNCR). The STANNCR is a component based mid-level representation, a base unit called "action component" is used to describe human actions. As illustrated in Figure 1(a), the action component is constructed by several correlated visual words, and the combination of action components forms the final representation for human action. Figure 1(b) gives a toy example for action video from class "walking". Learning from the visual words, we assume that the action "walking" includes two action components: arm movement and leg movement, then representation for walking is based on these two components. We adopt non-negative matrix factorization (NMF) for the action component learning and human action encoding. NMF decompose a non-negative matrix into two nonnegative matrices, the non-negative constraint keeps the component based property of representation. Meanwhile, to utilize the spatial-temporal information, we propose a Spatial Temporal Distribution Vector (STDV), which employs the Fisher vector and GMM to model the distribution of local feature locations corresponding to each visual word. The STDV is used as the part of graph regularization for NMF to incorporate spatial-temporal information for representation.

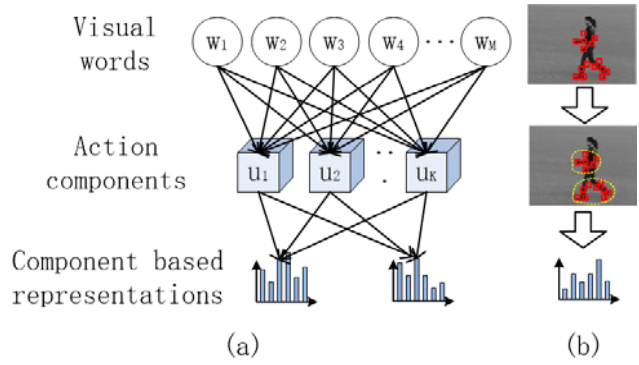

**Figure 1. Illustration of action component and component based representation. (a) Relationship among visual words, action components and component based representations. (b) A toy example for action video from class "walking".**

Figure 2 illustrates the flowchart of our proposal. Firstly, we extract local features from videos, and compute low-level representations and corresponding STDVs from local features. Then, for training data, we adopt ST-GNMF to compute the STANNCRs and action components simultaneously from the low-level representations and STDVS, and for testing sample, we use ST-GNMF to encode data with the action components and training data. Finally, SVM is applied for classification. The rest of this paper is organized as follows. Section 2 reviews the related work. Section 3 introduces the new spatial-temporal distribution vector (STDV). Section 4 presents the Spatial-temporal Aware Non-Negative Component Representation (STANNCR). Section 5 demonstrates the experimental results and comparisons. Section 6 concludes this paper.

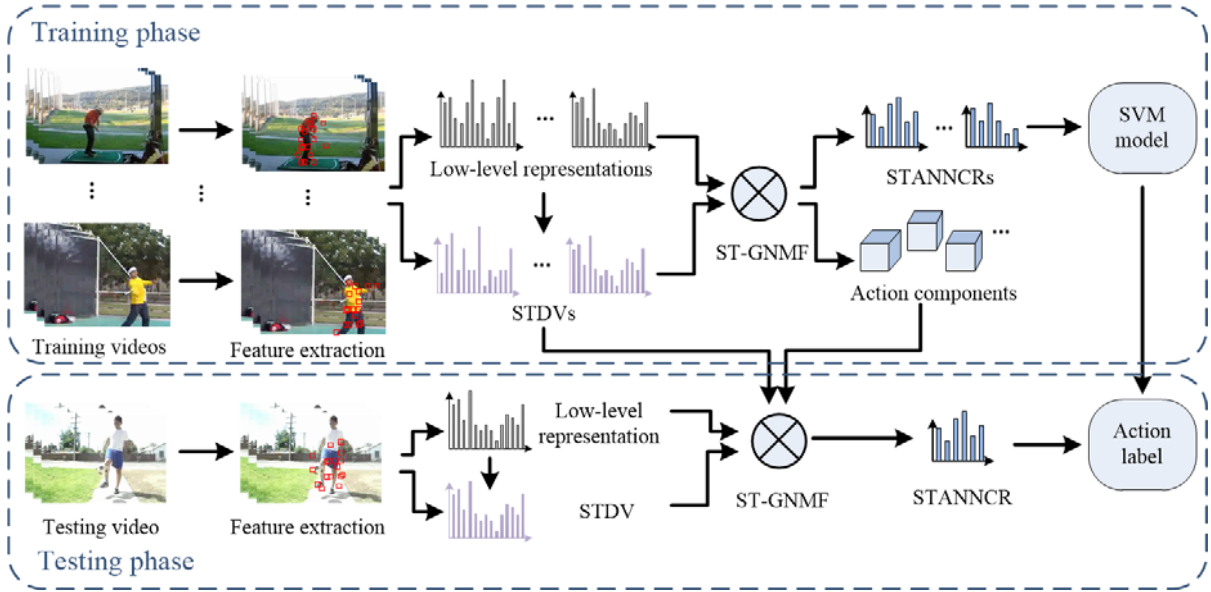

**Figure 2. Flowchart of the proposed work for action recognition.**

## 2. RELATED WORK

The low-level representation for visual recognitions is not discriminative enough, recently, researchers have exploited the mid-level representations derived from low-level features. Popular mid-level representation for visual recognitions include part-based model [10-15] and semantic attributes [16-19]. Han et al. [10] employ a cascade CRF to recognize the motion patterns for both the entire body and each body part in a learned hierarchical manifold space. Wang et al. [11] present a part based model to decompose an action into several parts to capture the local structure of the input data, and meanwhile they encoded pairwise relationships among different parts explicitly. Brenderl et al. [12] over-segment the whole video into tubes corresponding to action "part" and adopt spatial-temporal graphs to learn the relationship among the parts. Raptis et al. [13] group the trajectories into clusters with a graphical model, each of cluster can be seen as an action part. Wang et al. [14] develops motionlet to represent action video, where motionlet is spatial-temporal part with coherent appearance and motion features. In [15], the author propose to represent actions by a set of intermediate concepts called action units, the action units are learned from the training data with nonnegative matrix factorization, which lead to a part-based representation. For part based approaches, different methods have different definitions of part. The discrimination effectiveness of the parts in various situations is the key issues for part based method. In recent years, a semantic concept "attribute" is proposed to bridge the semantic gap between low-level features and high-level categories. Attribute based methods use human knowledge to create descriptors that capture intrinsic properties of actions. Liu et al. [16] explores both human-specified attribute and data-driven attribute classifiers to describe human actions by considering multiple semantic concepts. Parikh and Grauman [18] proposed relative attributes to capture more general semantic relationships which enable richer descriptions for images. Sadanand and Corso [17] show substantial improvements over standard benchmarks by using a bank of action detectors sampled broadly across semantic and viewpoint spaces. Li et al. [19] decompose a video into short segments, and use dynamics of attributes to characterize them. For most of attribute methods, the attributes need to be predefined and the attribute labels for training data require manual assignment by domain experts.

The BoVW framework ignores the spatial-temporal information, which is the important cue for action recognition. The dominant approach to incorporate spatial temporal information is the spatial-temporal pyramid (STP) [3], which is inspired by the spatial pyramid matching (SPM) [20] using in the image classification. STP partitions a video to a set of cells and yields the final representation by concatenating the histogram in each grid. Recent efforts [21-23] have tried to extent spatio-temporal pyramid by learning dynamic segmentation adapted to specific task. Cao et al. [21] present a scene adapted pooling. Ballas [22] and Nguyen [23] propose to segment videos dynamically based on video saliency. The high dimensionality of STP leads to computational infeasibility and huge storage consumption. Besides, STP can only obtain simple spatial and temporal information which is far from enough. Using spatial and temporal context as additional information to describe local features is another way to address this problem. Sun et al. [24] propose a hierarchical framework to encode point-level, intra-trajectory level, and inter-trajectory level spatio-temporal context information of video sequences. In [25], the author propose a representation that captures contextual interactions between interest points, based on the density of all features observed in each interest point's multiscale spatial-temporal contextual domain. Wang et al. [15] presents a locally weighted word context descriptor, encode each interest point by both itself and its neighborhood interest points. All these methods have benefits for action recognition, but they still stay in a relatively local level. As opposed to context method, recently some works [26, 27] use Fisher vector or super vector to model the global layout of local features. Krapac et al. [27] introduce the spatial Fisher vector to learn the location model by computing per visual word the mean and variance of spatial coordinates for corresponding patches. Yang and Tian [26] propose a super location vector to model the spatial-temporal locations of local features, which can be regarded as a simplified Fisher vector with sparse coding, and they report promising result for action recognition.

Our representation method is related to the work of Wang et al. [15]. We both utilize graph regularized NMF to learn action components from training data, and use the action components to represent the action videos. However, our proposal involves the spatial-temporal distribution information in the learning process to improve the discrimination of final representations. And we also use NMF to encode samples, which keeps the final results are nonnegative. For spatial-temporal information description, our method is relevant to [26] and [27]. The Fisher vector is adopt to model the distribution of locations assigned for each visual word. Unlike [26] and [27], we employ a novel fusion method for the spatial-temporal cues. We utilize the spatial-temporal distribution information as graph regularization for NMF, not simply concatenating the vector with other feature vectors.

## 3. SPATIO-TEMPORAL DISTRIBUTION VECTOR

In this section, we introduce spatial-temporal distribution vector (STDV) to utilize the spatial-temporal information. Consider the features encoded with the same visual word exhibit distinctive spatial temporal layout, the STDV is intended to capture this correlation between local features and the feature location distributions. We gather the features quantized to the same visual word, and employ Fisher vector and GMM to model the distribution of these local feature locations per visual word. The details of STDV is described as below.

Fisher vector records the deviation of data with respect to the parameters of a generative model. In recent evaluations [28, 29], it shows an improved performance over bag of features for both image and action classification. For the $k$-th visual word, we model the locations associated with it by a GMM distribution:

$$p(\mathcal{L}_k) = \sum_{g=1}^{G} \pi_{kg} \mathcal{N}(\mathcal{L}_k; \boldsymbol{\mu}_{kg}, \boldsymbol{\sigma}_{kg}), \quad (1)$$

where $\mathcal{L}_k = \{l_{k1},...,l_{kT_k}\}$ represents the locations of all local features represented by the $k$-th visual word, $\mathcal{L}_k \in \mathbb{R}^{3 \times T_k}$. $\pi_{kg}$, $\boldsymbol{\mu}_{kg}$ and $\boldsymbol{\sigma}_{kg}$ are the prior mode probability, mean vector and covariance matrix (assumed diagonal) of the $g$-th Gaussian model, respectively. Let $\boldsymbol{u}_{kg}$ and $\boldsymbol{v}_{kg}$ be the gradient of log-likelihood with respect to $\boldsymbol{\mu}_{kg}$ and $\boldsymbol{\sigma}_{kg}$ of the $g$-th Gaussian. $\gamma_i^{kg}$ denotes the soft assignment of $g$-th Gaussian component. After standard mathematical derivations and normalization, we obtain,

$$\boldsymbol{u}_{kg} = \frac{1}{T_k \sqrt{\pi_{kg}}} \sum_{i=1}^{T_k} \gamma_i^{kg} \left( \frac{l_{ki} - \boldsymbol{\mu}_{kg}}{\boldsymbol{\sigma}_{kg}} \right), \quad (2)$$

$$\boldsymbol{v}_{kg} = \frac{1}{T_k \sqrt{2\pi_{kg}}} \sum_{i=1}^{T_k} \gamma_i^{kg} \left[ \frac{(l_{ki} - \boldsymbol{\mu}_{kg})^2}{\boldsymbol{\sigma}_{kg}^2} - 1 \right]. \quad (3)$$

It's worth noting that when using soft assign method to encode local features, each local feature corresponding to more than one visual word with different weights. In other words, one location may belong to several different visual word, and each location in $\mathcal{L}_k$ has its own weight. Let $w_{ki}$ represents the weight for the corresponding location in $\mathcal{L}_k$. The Equation (2) and (3) should be revised for weighted locations as:

$$\boldsymbol{u}_{kg} = \frac{1}{\sum_{i=1}^{T} w_{ki} \sqrt{\pi_{kg}}} \sum_{i=1}^{T_k} w_{ki} \gamma_i^{kg} \left( \frac{l_{ki} - \boldsymbol{\mu}_{kg}}{\boldsymbol{\sigma}_{kg}} \right), \quad (4)$$

$$v_{kg} = \frac{1}{\sum_{i=1}^{T} w_{ki}\sqrt{2\pi_{kg}}} \sum_{i=1}^{T_k} w_{ki}\gamma_i^{kg} \left[ \frac{(l_{ki} - \mu_{kg})^2}{\sigma_{kg}^2} - 1 \right]. \tag{5}$$

The Fisher vector $x_K$ for $\mathcal{L}_k$ is the concatenation of $u_{kg}$ and $v_{kg}$ vectors for $g = 1,...,G$. The final vector representation $Z$ of STDV is the concatenation of $x_K$ from $K$ visual words:

$$\begin{aligned} Z &= \left[ z_1^T \cdots z_K^T \right]^T \\ &= \left[ u_{1_1}^T v_{1_1}^T \cdots u_{1_G}^T v_{1_G}^T \cdots u_{K_1}^T v_{K_1}^T \cdots u_{K_G}^T v_{K_G}^T \right]^T. \end{aligned} \tag{6}$$

## 4. SPATIAL-TEMPORAL AWARE NONNEGATIVE COMPONENT REPRESENTATION

### 4.1 Introduction of NMF

Non-negative matrix factorization (NMF) [30] is a matrix decomposition algorithm where a non-negative matrix is factorized into two nonnegative matrices. Usually, it offers dimension reduction by converting a data matrix to multiplication of two smaller matrices. Compared with other matrix factorization methods, NMF obtains a representation of data using non-negative constraints, which lead to a parts-based and intuitive representation of each input data. The superior property enables NMF to find applications in range fields such as document clustering [30], face recognition [31] and so on.

### 4.2 Principle of STANNCR

Graph regularized NMF (GNMF) [32] is an extension of NMF, which add a graph regularization to consider intrinsic geometrical and discriminative structure of the data space. Inspired by GNMF, we propose spatial-temporal aware GNMF (ST-GNMF) to involve spatial-temporal cues, which considers both low-level feature representation structure and feature location spatial-temporal distribution in the graph regularization of GNMF. And based on ST-GNMF, the spatial-temporal aware non-negative component representation (STANNCR) is presented for Action Recognition. Unlike previous works represent actions with low-level features, the STANNCR is a mid-level representation which extract action components from the low-level representation and encode videos with action components based on ST-GNMF. The ST-GNMF adds spatial-temporal distribution to the graph regularization. It is expected that if two video samples have the similar mid-level representations, they should have not only the similar intrinsic geometry of low-level feature representation, the corresponding spatial-temporal distributions should also be closed to each other. The proposed ST-GNMF is intended to minimize the object function as follows:

$$O = \|Y - UV\|^2 + \frac{\lambda}{2} \sum_{i,j=1}^{N} \|v_j - v_i\|^2 \left( \beta W_{ij}^F + (1-\beta) W_{ij}^D \right), \tag{7}$$

where $Y = [y_1, \cdots, y_N] \in \mathbb{R}^{M \times N}$, $y_i$ denote the $M$-dimensional low-level feature representation vector for the $i$-th video sample. $U = [u_1, \cdots, u_K] \in \mathbb{R}^{M \times K}$ and $V = [v_1, \cdots, v_N] \in \mathbb{R}^{K \times N}$ are two non-negative matrices. Consider each column of matrix $U$ as an action component constructed by several correlated visual words, then, $U$ becomes the action component dictionary, and each column of matrix $V$, denoted by $v_i$, is the new mid-level representation for the corresponding video sample based on the action component dictionary $U$.

The second part of objective function is the newly added graph regularization. The tradeoff parameter $\beta$ controls the impact of spatial-temporal distribution to the object function, when $\beta = 1$, the ST-GNMF degenerate to the standard GNMF. $W^F \in \mathbb{R}^{N \times N}, W^D \in \mathbb{R}^{N \times N}$ represent the weight matrix for low-level feature representation and spatial-temporal distribution vector respectively. We adopt the heat kernel weight for both $W^F$ and $W^D$, which is define as:

$$W_{ij}^F = \exp\left( -\frac{1}{\delta} \|y_j - y_i\|^2 \right), \tag{8}$$

$$W_{ij}^D = \exp\left( -\frac{1}{\delta} \|z_j - z_i\|^2 \right), \tag{9}$$

where $z_i$ represents STDV for the $i$-th video sample based on $y_j$.

### 4.3 Implementation for STANNCR

Define $W = \beta W^F + (1-\beta) W^D$ and $D_{ii} = \sum_j W_{ij}$, then we can obtain the Laplacian matrix $L = D - W$, and the object function can be simplified to:

$$O = \|Y - UV\|^2 + \lambda Trace(VLV^T). \tag{10}$$

This is the same form as GNMF, and it can optimized with the same method for GNMF. The object function is not convex in both $U$ and $V$ together, but it is convex in $U$ only or $V$ only. Following the work [32], we optimize $U$ and $V$ alternatively using two iterative update algorithms. The updating rule is as follows:

$$U \leftarrow U \odot \frac{YV^T}{UVV^T}, \tag{11}$$

$$V \leftarrow V \odot \frac{U^T Y + \lambda VW}{U^T UV + \lambda VD}, \quad (12)$$

Where $\odot$ is an element-wise product and all divisions in (11) and (12) are element-wise divisions.

For testing videos, we first extract low-level representation $y_{ti}$ and the spatial-temporal distribution $z_{ti}$ for each testing video. Define $Y_t = [y_{t1}, \cdots, y_{tN_t}]$ and $Z_t = [z_{t1}, \cdots, z_{tN_t}]$, $N_t$ is the number of testing videos. The simplest way to compute the mid-level representation matrix is $V_t = U^\dagger Y_t$, where † represents the Moore-Penrose pseudoinverse. However in such a case, $V_t$ might have negative elements, moreover, it could not take advantage of spatial-temporal cues in $Z_t$. To keep consistent with the training data representation, we still use ST-GNMF to solve $V_t$. Consider a cost function $O_t$ similar as Equation (7) with both training and testing data,

$$\begin{aligned} O_t &= \left\| \hat{Y} - U\hat{V} \right\|^2 + \frac{\lambda}{2} \sum_{i,j=1}^{N+N_t} \left\| v_j - v_i \right\|^2 \left( \beta \hat{W}_{ij}^F + (1-\beta)\hat{W}_{ij}^D \right) \\ &= \left\| \hat{Y} - U\hat{V} \right\|^2 + \lambda Trace\left( \hat{V} \hat{L} \hat{V}^T \right) \\ &= \left\| Y - UV \right\|^2 + \left\| Y_t - UV_t \right\|^2 + \lambda Trace\left( \hat{V} \hat{L} \hat{V}^T \right). \end{aligned} \quad (13)$$

The symbol ˆ indicates the matrix contains data for both training and testing samples. Fixing $U$ and $V$, $V_t$ can be learned by minimizing object function $O_t$. Taking derivatives with respect to $V_t$, and define $\hat{L} = \begin{bmatrix} L & L_2 \\ L_2^T & 0 \end{bmatrix}$, we have

$$\frac{\partial O_t}{\partial V_t} = -2U^T Y_t + 2U^T UV + 2\lambda VL_2, \quad (14)$$

From Equation (14), we can get the updating rule for $V_t$ as follow:

$$V_t \leftarrow V_t \odot \frac{U^T Y_t + \lambda VW_2}{U^T UV_t^T + \lambda VD_2}, \quad (15)$$

where $W_2$ and $D_2$ are the corresponding sub-matrices to $L_2$. Equation (15) shows that $V_t$ relies on not only the action component dictionary $U$ but also the training data, the encoding algorithm for testing videos keeps the consistency between training and testing representations.

The outline of our proposed STANNCR is summarized in algorithm 1. And the encoding method for new testing videos is listed in algorithm 2.

---

**Algorithm 1** Computation of STANNCR

**Input**: training videos $X = \{x_i\}$

**Output**: action component dictionary $U$ and STANNCRs $V = [v_1, \cdots, v_N]$ for $X$

Compute low-level feature representations $Y = [y_1, \cdots, y_N]$ and STDVs $Z = [z_1, \cdots, z_N]$ for $X$

Compute weight matrices $W^F$ and $W^D$ using $Y$ and $Z$ according to (8) and (9)

Compute $W$, $D$ and Laplacian Matrix $L$ using $W^F$ and $W^D$

Initialize $U$ and $V$ as nonnegative random matrices

**while** not converge **do**
  Fix $V$, Update $U$ according to (11)
  Fix $U$, Update $V$ according to (12)
**end while**

> **Algorithm 2** Encoding method for testing videos
>
> **Input**: testing videos $X_t = \{x_{ti}\}$, action component dictionary $U$, and $Y, Z, V$ for training videos $X$
>
> **Output**: STANNCRs $V_t$ for $X_t$
>
> Compute low-level feature representations $Y_t = [y_{t1}, \cdots, y_{tN_t}]$ and STDVs $Z_t = [z_{t1}, \cdots, z_{tN_t}]$ for $X_t$
>
> Compute $\hat{Y} = [Y, Y_t]$, $\hat{Z} = [Z, Z_t]$
>
> Compute weight matrices $\hat{W}^F$ and $\hat{W}^D$ using $\hat{Y}$ and $\hat{Z}$ according to (8) and (9)
>
> Compute $\hat{W}$, $\hat{D}$ and Laplacian Matrix $\hat{L}$ using $\hat{W}^F$ and $\hat{W}^D$
>
> Initialize $V_t$ as nonnegative random matrix
>
> **while** not converge **do**
>   Fix $U$ and $V$, Update $V_t$ according to (15)
>
> **end while**

## 4.4 Advantages of STANNCR

There are several advantages of the proposed spatial-temporal aware non-negative component representation.

1. We propose the ST-GNMF to learn the non-negative representations for both training and testing videos, which leads to component-based representations for all samples. The action component considers the correlations between visual words, compared with BoVW model, it reduces the dimension of representation and makes the representation more discriminative.

2. The spatial-temporal distribution is taken into account in both training and testing phase, which makes the representation more discriminative than appearance only method. Especially for those classes with similar appearance based representations. Moreover we use the distribution as the regularization constraint for GNMF, not simply concatenating the distribution vector with the corresponding data vector. Our fusion method makes the final representation vector simple and efficient.

3. The encoding method for testing videos considers the correlations between training and testing data. The representations for testing samples are learned with not only the action components but also the training data, which keeps the consistency of encoding between training and testing videos.

## 5. EXPERIMENTAL RESULTS

### 5.1 Dataset and setup

*5.1.1 Dataset*

We extensively evaluate the proposed method on three popular human action datasets: KTH, YouTube and HMDB51. Some sample frames from these datasets are illustrated in Figure (3). The experimental settings of these datasets are summarized as follows:

The KTH dataset [33] contains of six human action classes: walking, jogging, running, boxing, waving and clapping. Each action is performed several times by 25 subjects. The sequences were recorded in four different scenarios: outdoors, outdoors with scale variation, outdoors with different clothes and indoors. The background is homogeneous and static in most sequences. In total, the data consists of 2,391 video samples. We follow the experimental settings in [33] where samples are divided into the training set (16 subjects) and the testing set (9 subjects).

The YouTube dataset [34] contains 11 action categories: basketball shooting, biking/cycling, diving, golf swinging, horseback riding, soccer juggling, swinging, tennis swinging, trampoline jumping, volleyball spiking, and walking with a dog. This dataset is challenging due to large variations in camera motion, object appearance and pose, object scale, viewpoint, cluttered background and illumination conditions. The dataset contains a total of 1,168 sequences. Following the original setup [34], we use Leave-One-Group-Out cross-validation and report the average class accuracy.

The HMDB51 dataset [35] is a large action video database with 51 action categories and 6,766 video sequences which are collected from a variety of sources ranging from digitized movies to YouTube website videos. HMDB51 contains facial actions, general body movements and human interactions. It is a very challenging benchmark due to its high intra-class variation and other fundamental difficulties. We follow the experimental settings in [11] where three train-test splits are available, and we report average accuracy over the three splits.

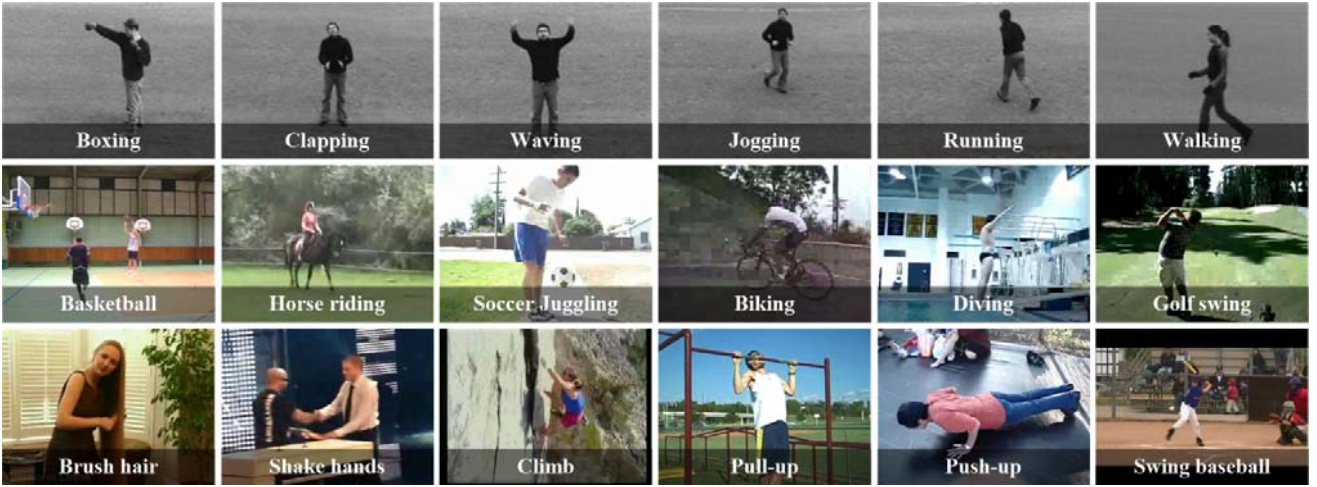

**Figure 3. Sample frames from the three action recognition datasets used in our experiments. From *top to bottom* KTH, YouTube and HMDB51 datasets.**

*5.1.2 Low-level features setting*

Considering the success of dense sampling in image classification and action recognition, we evaluate our approach on three features based the dense trajectory [6]: HOG, HOF, and motion boundary histogram (MBH). HOG focuses on static appearance cues, whereas HOF captures local motion information. MBH computes gradient orientation histograms from horizontal and vertical spatial derivatives of optical flow. It has been proven effective to represent motion information and suppress camera motion. So for each action video clip, we compute three features: HOG (96), HOF (108), and MBH (192), where the number in parentheses denotes the descriptor dimensionality.

For all experiments, we use the same dictionary for each feature, and the dictionary size is set to 2000. We employ localized soft assignment [7] for low-level representation encoding. Localized soft assignment has better accuracy than vector quantization, and can keep the encoding results non-negative, which is important for the further process.

*5.1.3 Classification setting*

For classification we employ a non-linear SVM with an RBF-$\chi^2$ kernel, given two video STANNCRs $v_i$ and $v_j$, the RBF-$\chi^2$ kernel is defined as:

$$K(v_i, v_j) = \exp\left(-\frac{1}{A} D(v_i, v_j)\right) \quad (16)$$

$$D(v_i, v_j) = \frac{1}{2} \sum_k \left( \frac{(v_i^k - v_j^k)^2}{v_i^k + v_j^k} \right) \quad (17)$$

where $D(v_i, v_j)$ is the $\chi^2$ distance between $v_i$ and $v_j$. $A$ is the average value of all the distances in training samples. As action recognition is a multi-class classification problem, we use a one-against-rest strategy and select the class with the highest score.

## 5.2 Evaluation of STDV

In our approach, the STDV is employed to construct regularization constraint of ST-GNMF. Before experiments for ST-NMF, we evaluate the effectiveness of STDV first. The STDV is compared to the widely used spatial-temporal pyramid (STP) and spatial-temporal location Fisher vector (STLFV). STLFV can be regarded as the STDV without spatial scale normalization. For STP, four different spatio-temporal segments are used in our experiments. We apply a 1 × 1 whole spatial block and a 2 × 2 spatial grid for the spatial domain. For the temporal domain, the entire sequence and two temporal segments are employed. The combination of these subdivisions in both spatial and temporal domains generates 15 space-time cells in total. The final representation vector for STP is the concatenation of low level representations for each cell. For STDV and STLFV, the final vector is the combination of low level representation and spatial-temporal representation. For the fair comparison, we use the same visual vocabulary for all three methods and the same additional location dictionary for STDV and STLFV. The size of visual vocabulary size is set to 2000, and the location dictionary size is 5. The comparison result is listed in Table 1.

**Table 1. Comparison of STDV and other method for modeling spatial-temporal information for a variety of features on the HMDB51 dataset**

|  | HOG | HOF | MBH |
|---|---|---|---|
| none | 34.8% | 39.5% | 45.2% |
| STP | 36.2% | 40.4% | 46.5% |
| STLFV | 37.6% | 41.9% | 47.8% |
| STDV | **38.1%** | **42.3%** | **48.2%** |

As shown in table 1, all of three method can improve the results, because of spatial-temporal information complemented to the appearance and motion representations. However, STLFV and STDV achieve more significant improvement for all features, and the representation vectors are more compact than STP. In our experiment, the dimensions of STP is $15 \times m \times 2000$, and the dimensions for STLFV and STDV is $(15+m) \times 2000 \times 2$, where $m$ is the descriptor dimension. Moreover, as spatial scale normalization is considered, the proposed STDV has higher accuracies than STLFV with the same vector dimensionality.

## 5.3 Evaluation of STANNCR

### 5.3.1 Comparison with BoVW and GNMF based representation

The STANNCR is compared with Bag of Visual Words (BoVW) representation and GNMF based mid-level representation on three datasets. BoVW is the most popular method in recent years, and the proposed STANNCR is based on the BoVW result. GNMF based representation is similar to STANNCR, expect that STANNCR adds the spatial-temporal distribution to regularization constraint. HOG, HOF and MBH are adopted as low level features. All three methods use the same visual dictionary and same settings for each low-level feature. For STANNCR, we set the tradeoff parameter $\beta = 0.6$.

Table 2, Table 3 and Table 4 report the comparison results on KTH, YouTube and HMDB51 datasets respectively. We can observe that GNMF based method and STANNCR significantly outperform the results of BoVW on three datasets. GNMF based representation and STANNCR are both mid-level methods using non-negative component for representation. Compared with BoVW, STANNCR has 3.74%, 5.9% and 5.84% average improvements on KTH, YouTube and HMDB51 dataset respectively, and the average improvements for GNMF are 2.1%, 2.93% and 3.67%. Another comparison is between GNMF based representation and STANNCR, by adding spatial-temporal distribution as regularization constraint, STANNCR achieves higher accuracies than GNMF based method. The average improvements are respectively 1.64%, 2.97% and 2.17% on three datasets. The study on three representation methods demonstrates the effectiveness of the non-negative component representation and the spatial-temporal distribution information for classification.

**Table 2. Comparison of STANNCR and other representation methods with different low-level features on KTH**

|  | HOG | HOF | MBH |
|---|---|---|---|
| BoVW | 87.5% | 93.1% | 94.6% |
| GNMF | 89.8% | 95.4% | 96.3% |
| STANNCR | **92.3%** | **96.3%** | **97.6%** |

**Table 3. Comparison of STANNCR and other representation methods with different low-level features on YouTube**

|  | HOG | HOF | MBH |
|---|---|---|---|
| BoVW | 73.9% | 71.2% | 81.7% |
| GNMF | 77.1% | 73.7% | 84.8% |
| STANNCR | **80.2%** | **76.9%** | **87.1%** |

**Table 4. Comparison of STANNCR and other representation methods with different low-level features on HMDB51**

|  | HOG | HOF | MBH |
|---|---|---|---|
| BoVW | 37.3% | 39.5% | 46.4% |
| GNMF | 38.6% | 43.4% | 48.5% |
| STANNCR | **40.2%** | **44.8%** | **51.3%** |

### 5.3.2 Performance for combining all features

To further elevate the recognition accuracy, we try to combine STANNCRs with all features before classification, the result is illustrated in Table 5. HMDB51 and YouTube dataset obtain 3.9% and 2.1% extra improvements. The result on KTH dataset has no improvement, as the single feature result is already close to 100% and the remaining error samples are hard to classify, it's difficult to further increase the recognition accuracy.

**Table 5. Recognition accuracies of combining STANNCRs with all features on difference datasets**

|  | KTH | YouTube | HMDB51 |
|---|---|---|---|
| Accuracy | 97.8% | 89.2% | 55.4% |

### 5.3.3 Effects of encoding method for component based representation

In our proposal, the training samples encoding and action components learning are done simultaneously by ST-GNMF. And the representation vectors for testing samples are computed by ST-GNMF with a fixed action component dictionary. Besides our method, work [36] and [15] also apply Locality-constrained Linear Coding (LLC) and Sparse Coding (SC) for encoding samples with non-negative action component. We conduct experiment on HMDB51 dataset to study the effects of these three encoding methods.

As shown in Figure 4, for all three features, our method yields better accuracies than LLC and SC. We analyze three reasons for this result. First, our encoding method considers spatial-temporal distribution as constraint in both training and testing samples, while LLC and SC totally ignore spatial-temporal information. Second, our method guarantees the encoding results are non-negative, which keeps the part based property for the final representation. LLC and SC may have negative elements in the codes. Third, our method encodes the testing videos with not only the action components but also the training data, which keeps the consistency of encoding between training and testing videos, while the other two methods only consider the action components for testing data encoding.

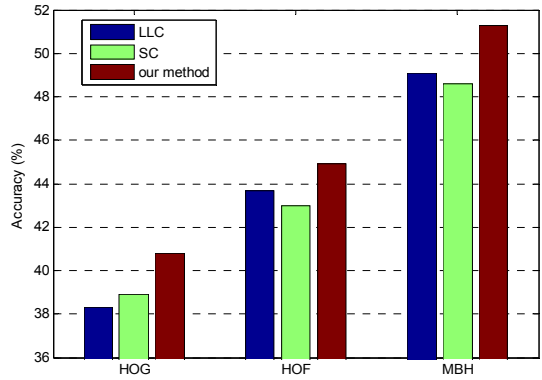

**Figure 4. Comparison of different encoding method for component based representation on HMDB51**

### 5.3.4 Effects of tradeoff parameter

We discuss the variation of the tradeoff parameter $\beta$ to the performance of STANNCR. The feature HOG is used for evaluation, and the experiments are conducted on KTH and YouTube datasets. Figure 5 demonstrate how the performance varies with the parameter $\beta$.

As we can see, for the tradeoff parameter $\beta$, the performance reach peak around $\beta = 0.6$, then the accuracy decreases on both sizes. $\beta$ controls the impact percentage of STDV, this shows that motion or appearance information and spatial-temporal cues are mutually complementary, combining both them can achieve better result than only using one. We set $\beta = 0.6$ for all the experiments.

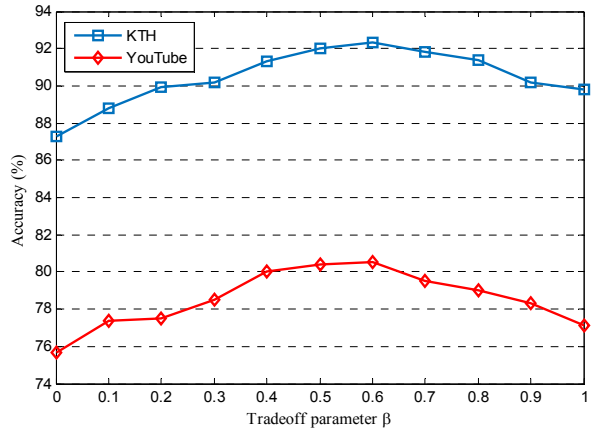

**Figure 5. The performance of STANNCR for a variety of tradeoff parameter $\beta$ on KTH and YouTube**

## 5.4 Comparison with State-of-the-Art Results

Table 6 presents the comparison between our best results and several recent results reported in the literature for all three datasets. We can observe that our method outperforms the state-of-the-art results on YouTube and HMDB51 datasets, and the result on KTH dataset is also comparable to the best reported results. The improvement over the best reported result is 1.5% on the HMDB51 dataset, and 1.2% on the YouTube dataset.

Table 6. Comparison with the state-of-the-art results

| KTH | | YouTube | | HMDB51 | |
|---|---|---|---|---|---|
| Approach | Accuracy | Approach | Accuracy | Approach | Accuracy |
| Le et al. [37] | 93.9% | Le et al. [37] | 75.8% | Sadanand et al. [17] | 26.9% |
| Sadanand et al. [17] | 98.2% | Bhattacharya et al. [38] | 76.5% | Jiang et al. [39] | 40.7% |
| Ji et al. [40] | 90.2% | Wang et al. [6] | 85.4% | Shi et al. [41] | 47.6% |
| Wang et al. [6] | 95.3% | Wang et al. [15] | 82.2% | Wang et al. [6] | 48.3% |
| Wang et al. [15] | 95.5% | Yang et al. [26] | 88.0% | Yang et al. [26] | 53.9% |
| Our Method | **97.8%** | Our Method | **89.2%** | Our Method | **55.4%** |

## 6. CONCLUSION

In this paper, we have presented a novel mid-level representation for action recognition. The proposed STANNCR is based on action component and considers the spatial-temporal information. An effective STDV is first introduced to model the spatial-temporal distributions in a compact and discriminative manner. Then, a novel ST-GNMF is proposed to learn the action components and encode the video samples with the action components. The ST-GNMF adopts STDV as graph regularization constraint to incorporate the spatial-temporal cues for final representation. Our approach has been extensively tested on three datasets, the result of experiments demonstrates the effectiveness of STANNCR for action recognition.

## 7. CONFLICT OF INTERESTS

The authors declare that there is no conflict of interests regarding the publication of this paper.